%% file: root.tex
\documentclass[journal]{IEEEtran}

\usepackage{amssymb,amsmath}

\usepackage{amsthm}
\usepackage{mathtools}
\usepackage{cancel}
\usepackage{multirow}
\usepackage{graphicx}
\usepackage{comment}
\usepackage[sort,compress,noadjust]{cite}
\usepackage[font=footnotesize]{subcaption}
\usepackage[font=footnotesize]{caption}
\usepackage{booktabs}
\usepackage{makecell}

\usepackage{amsfonts}
\usepackage{tabularx}
\usepackage{graphicx}
\usepackage{array}
\usepackage{float}
\usepackage{hhline}
\usepackage{algorithm}
\usepackage{algorithmic}
\usepackage{float}
\usepackage[colorlinks,urlcolor=blue]{hyperref}
\usepackage{setspace}
\usepackage{colortbl}
\usepackage[dvipsnames]{xcolor}
\usepackage[normalem]{ulem}
\usepackage{comment}
\usepackage{arydshln}
\usepackage{accents}

\usepackage{placeins}

\theoremstyle{plain}

\theoremstyle{plain}

\theoremstyle{plain}

\theoremstyle{plain}

\theoremstyle{definition}

\theoremstyle{definition}

\theoremstyle{remark}

\usepackage[capitalize]{cleveref}
\crefformat{equation}{(#2#1#3)}
\Crefformat{equation}{Equation~(#2#1#3)}
\Crefname{equation}{Equation}{Eqs.}
\crefname{Structure}{Struct.}{Structs.}
\Crefname{Structure}{Structure}{Structures}
\creflabelformat{Structure}{#2#1#3}

\usepackage{censor}
\StopCensoring   

\title{\LARGE \bf
Online Geometric Change Detection via Scene Decomposition
}

\author{\censor{David Thorne$^{1}$}, \censor{Samuel Jia Cong Chua$^{1}$}, \censor{Nakul Joshi$^{1}$}, \censor{Aiden Wong$^{1}$}, \\\censor{Christa S. Robison$^{2}$}, \censor{Philip Osteen$^{2}$}, \censor{Brett T. Lopez$^{1}$}%
\thanks{\xblackout{*This research was sponsored by the DEVCOM Army Research Laboratory (ARL) under SARA CRA
W911NF-24-2-0017. Distribution Statement A: Approved for public release; distribution is unlimited.}}
\thanks{\xblackout{$^{1}$University of California, Los Angeles,  Los Angeles, CA, USA }{\tt\small\xblackout{\{davidthorne, samuelchua, nakuljoshi, aidenwong, btlopez\} @ucla.edu}}}%
\thanks{\xblackout{$^{2}$DEVCOM Army Research Laboratory (ARL), Adelphi, MD, USA. \{\texttt{christa.s.robison, philip.r.osteen \}.civ@army.mil}}}%
}%

\begin{document}

\maketitle
\addtolength{\topmargin}{0.05in}

\thispagestyle{empty}
\pagestyle{empty}

\begin{abstract}

Autonomous robots are increasingly deployed on long duration single- and multi-session missions in dynamic environments, where the ability to identify environmental changes such as fallen trees or opened doors provides important contextual information for online planning.
We propose a framework called Change Detection via Scene Decomposition (CDSD) for accurate online geometric change detection using LiDAR or RGB-D sensors.
Recent advances in geometric SLAM have made it possible to generate dense, tightly aligned maps without post processing, but comparing global maps across entire sessions is computationally expensive and does not allow for single-session online change detection.
CDSD instead spatially decomposes mapped environments into unique scenes where changes can be found efficiently by comparing dense, local subsets of the global map called submaps.
As the first submap-based approach for geometric change detection, we identify and address the following core challenges: 1) identifying appropriate scenes for change detection that require minimal redundant information; 2) generating dense and representative submaps for each scene; 3) detecting changes between submaps with differing fields of view; and 4) processing detected changes for real-time map reconstruction.
Results demonstrate our algorithm on custom datasets collected at the \xblackout{Army Research Laboratory facility in Graces Quarters, Maryland}, and on open-source multi-session change detection datasets. 

\end{abstract}


\input{sections/introduction.tex} 
\input{sections/related_works.tex}
\input{sections/methods.tex}
\input{sections/results.tex}
\input{sections/conclusion.tex}

\IEEEpubidadjcol

\bibliographystyle{IEEEtran}
\bibliography{references}

\newpage
\vfill

\end{document}

%% file: sections/introduction.tex
\section{Introduction}
\label{sec:introduction}

Geometric change detection identifies persistent changes in an environment, such as a fallen tree or opened door, by comparing point cloud maps acquired at different times.
Detecting changes online (i.e., within a few seconds of observing the change) enables real-time situational awareness that can be used by autonomous and human agents operating in dynamic environments \cite{krawciw2023changeofscenery, tranzatto2022cerberus}.
Achieving accurate online change detection is inherently difficult though as it requires dense, precisely aligned point clouds to detect small objects.
Recent LiDAR odometry and SLAM algorithms can provide centimeter accuracy \cite{chen2023direct, campos2021orb}, enabling online scan-to-scan comparisons without the need for downstream refinement.
However, individual scans concentrate points along a few scan lines that grow widely spaced with range, leaving them too sparse for reliable change detection.
This sparsity can be addressed by aggregating multiple viewpoints in a submap, which is the union of a selected subset of the scans collected within a bounded region.
Submaps recover the density for accurate change detection, but raise several open questions about how they should be constructed and compared online.

\begin{figure}[t!]
    \centering
    \includegraphics[width=0.48\textwidth]{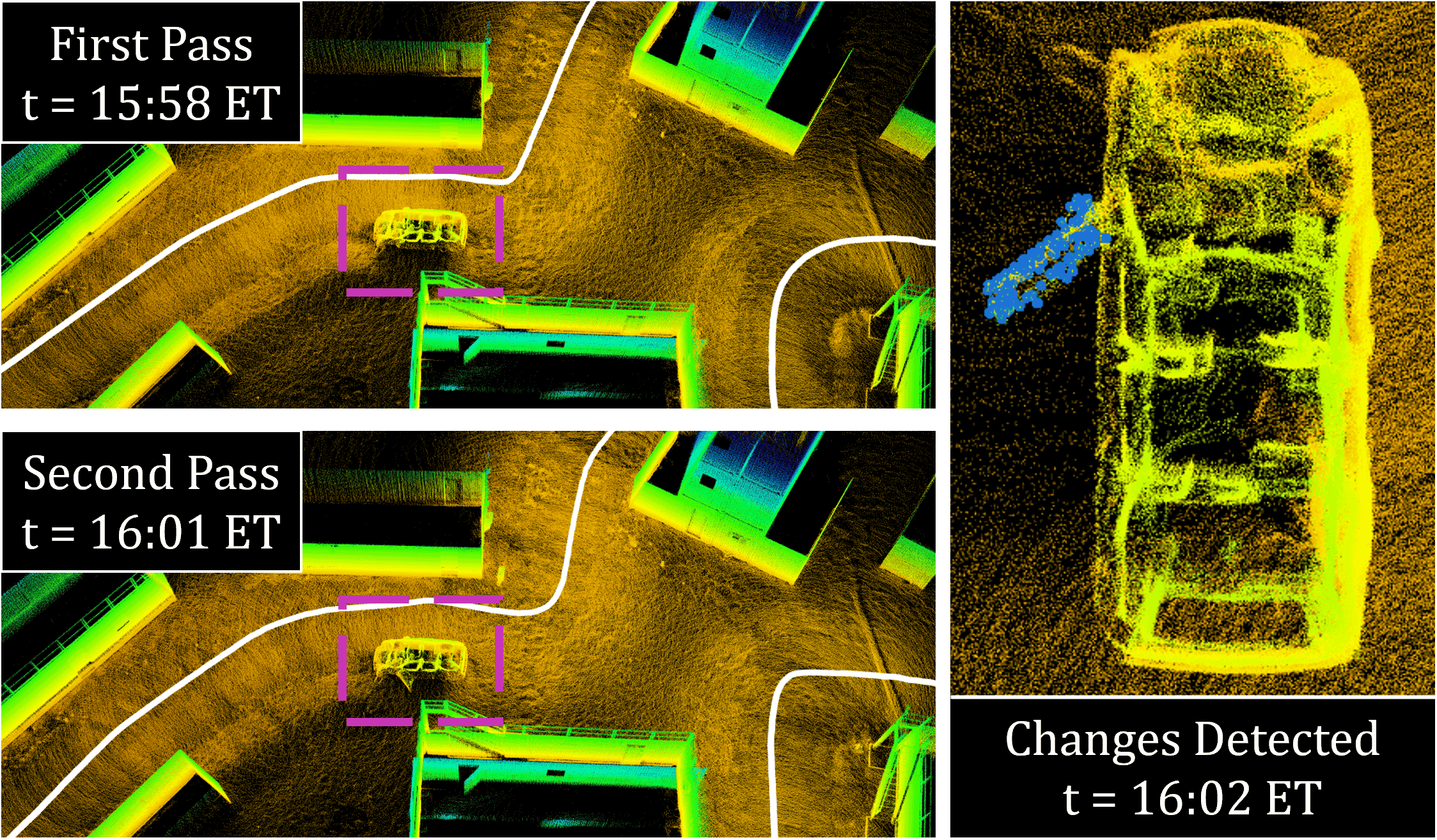}
    \caption{A car door is opened between the first and second mapping passes of the same session and detected in under one minute. Base points are colored by height (z-axis); change points are shown in blue in the right figure.}
    \label{fig:change_detection}
\end{figure}

The first geometric change detection methods compare global maps collected via airborne LiDAR surveys to monitor structural changes such as new or demolished buildings \cite{lague2013accurate}.
More recent approaches extend these ideas to ground-based mobile robotics, which use dynamic object removal and pose graph refinement to improve global map consistency for detecting small changes \cite{kim2022lt, gil2025ephemerality}.
While these methods achieve high accuracy, they are incompatible with real-time decision making because they require complete before and after global maps that are only available after a completed mapping session.
Current online geometric change detection methods instead compare individual scans against a prebuilt map to significantly reduce computation time per change detection instance \cite{wellhausen2017reliable}.
In addition to the limitation of using sparse individual scans, a prebuilt comparison map must either be produced offline from a previous session, or maintained online as a dense global map. 
Both prior maps waste computation time on non-overlapping regions and the latter grows unbounded in memory.

This paper proposes Change Detection via Scene Decomposition (CDSD), a framework that 1) decomposes the mapped environment into locally consistent regions called \emph{scenes}, 2) constructs representative cumulative and recent submaps for each scene immediately after it is mapped, and 3) detects changes between submaps despite possibly mismatched fields of view.
The resulting framework is both memory and computationally efficient as it only requires loading and operating on a concise set of relevant scans at any given time.
This paper addresses several unique challenges introduced by submap-based geometric change detection, where the novel contributions are:
\begin{itemize}
    \item \textbf{Scene Identification.} 
    The mapped environment is spatially decomposed into unique and non-overlapping scenes by an online scene identification algorithm.
    \item \textbf{Submap Generation.} 
    OptMap~\cite{thorne2025optmap}, a recently developed geometric map distillation algorithm, is used to generate representative, size-constrained submaps for each scene.
    \item \textbf{Change Detection.} 
    A novel submap-to-submap change detection algorithm identifies candidate changes and verifies they are not artifacts of submap pose, field of view, or occlusion differences.
    \item \textbf{Change Manager.} 
    An efficient change management system that leverages a lightweight data structure to provide a convenient interface for mission/behavior planners.
\end{itemize}
Our experimental results validate each of the four contributions through an ablation study, in which the complete framework detects 97.8\% of ground-truth changes across multiple custom datasets.
Removing any single component of the framework degrades detection performance or computation time.
The overall framework performance is then quantified by comparison against a state-of-the-art offline geometric change detection algorithm \cite{gil2025ephemerality}, where our framework reports change points that are consistently corroborated by the offline method.
CDSD is released as open-source code\footnote{\href{https://github.com/vectr-ucla/geometric_change_detection}{https://github.com/vectr-ucla/geometric\_change\_detection}} and validated with experimental results collected at the Army Research Laboratory facility in Graces Quarters, Maryland. 

\emph{Notation.} A point cloud is a finite set of points $P = \{\mathbf{p}_i \in \mathbb{R}^3\}_{i=1}^{N} \subset \mathcal{P}$ where $\mathcal{P}$ is the set of all possible point clouds, the 3D position of a scan is $\mathbf{x} \in \mathbb{R}^{3}$, and a descriptor\footnote{Descriptors are generated by a learned mapping from point clouds to feature space $\phi:\mathcal{P} \rightarrow \mathbb{R}^{d}$, where $\mathbf{d}=\phi(P)$ and the network is trained such that the distance between descriptors is monotonically decreasing with the overlap between input clouds.} is a fixed-length vector $\mathbf{d} \in \mathbb{R}^{d}$.
For every point cloud $P_k$, we have the tuple $(\mathbf{d}_{k}, \mathbf{x}_{k}, t_{k}) \in \mathfrak{T}$ where $t_{k}$ is the timestamp when the point cloud was acquired.
A scene is a bounded region denoted as $S \subset \mathbb{R}^{3}$, with volume $|S|$, and the set of all scenes is $\mathfrak{S}=\{S_{1},S_{2},...\}$.
We denote the set of descriptors collected within a scene $S$ as $D_S = \{ \mathbf{d} \in \mathbb{R}^d \, | \, \forall (\mathbf{d}, \mathbf{x}, t) \in \mathfrak{T}, \, \mathbf{x} \in S \}$.
Each change detection instance is defined for a scene $S_{j}$, which has been visited $v$ times, where $v \geq 2$.
For each instance, two submaps are used.
A cumulative submap $\mathcal{C}_{j}^{1:v-1}$ is a subset of the scans collected during visits $1$ to $v-1$, and a recent submap $\mathcal{S}_{j}^{v}$ is a subset of the scans collected during visit $v$.

%% file: sections/related_works.tex
\section{Related Works}
\label{sec:related_works}

Change detection for mobile robots originated with vision-based methods \cite{goyette2012changedetection}.
Recent visual change detection methods use increasingly sophisticated approaches to overcome challenges with varying points of view and environmental conditions.
Lin et al. \cite{lin2025robust} use cross-attention modules to compare multiple images of the same scene with varied fields of view, and more recently NeRFs and Gaussian splats have become popular for overcoming field of view differences \cite{martinson2024meaningful}.
These representations render the scene from a common viewpoint so that images captured along different trajectories can be compared directly.
Despite this progress, purely visual change detection still relies on training robust models and recognizing features from different viewpoints and with variable lighting.

Geometric change detection has a rich history in remote sensing \cite{lague2013accurate}, where aerial mapping is used to monitor environmental changes over time.
Within robotics, these methods can be separated into online and offline approaches.
Offline methods focus on multi-session environmental monitoring and static map building, where tasks such as dynamic object removal \cite{kim2020remove}, multi-session map alignment \cite{herraez2026multi, ali2026probper}, and dense change detection can be performed after mapping \cite{kim2022lt, gil2025ephemerality, rowell2024lista, lee2025lidar}.
Online geometric change detection has historically received less attention, but has become more important as autonomous robots are increasingly deployed into dynamic environments.
Methods that only use geometric sensing primarily perform scan-to-map change detection, comparing single scans to prior maps with distance-based thresholding \cite{qian2022pocd, wellhausen2017reliable, jang2026chamelion} or by converting LiDAR point clouds into range images and applying established vision-based change detection methods \cite{krawciw2024lasersam}.
Because a single scan is sparse and observes the scene from one viewpoint, these comparisons must separate genuine changes from differences in coverage and occlusion, which limits the density of changes that can be reported reliably.

Recent developments in visual foundation models have led to open-set image-text embeddings \cite{radford2021learning} and object detection and segmentation \cite{liu2024grounding, ravi2025sam}.
This allows robots equipped with geometric and visual sensing modalities to build spatio-temporal maps with embedded semantic knowledge.
Tracking semantic entities across sessions allows changes to be reported at the level of objects rather than individual points \cite{zhao2026supermap}.
While promising, this approach relies on a predefined vocabulary to query the map, which the authors of \cite{zhao2026supermap} note leaves the open-set problem unresolved, and requires multiple mutually calibrated sensing modalities rather than a single sensor.

%% file: sections/methods.tex
\section{Methods}
\label{sec:methods}

This paper proposes a framework for fast and accurate geometric change detection.
We assume each scan has an accurate odometry estimate and a learned point cloud descriptor \cite{ma2022overlaptransformer}.
An online map merging algorithm such as \cite{ma2026pinnet} additionally allows the framework to handle multi-session change detection.
At its core, the framework compares concise, dense submaps from unique locations to strike a balance between computation time and retaining sufficient geometric detail.
The central insight is the spatial decomposition of the mapped environment into scenes, effectively splitting the large offline problem into several, much smaller subproblems with minimal effect on accuracy.
Scene-based change detection requires addressing several subproblems, which we outline here.
In \cref{subsec:scenes}, an incremental algorithm is introduced to identify unique, non-overlapping scenes.
\Cref{subsec:submap_gen} then describes how near-optimal cumulative and recent submaps $\mathcal{C}_{j}^{1:v-1}$ and $\mathcal{S}_{j}^{v}$ are selected for each scene.
Next, \cref{subsec:change_detect_alg} presents a submap-to-submap change detection algorithm with multiple filters which account for differences in submap positions and fields of view.
Finally, in \cref{subsec:change_manager}, change points are clustered into change objects, which simplify map maintenance and provide actionable data for planners.

\subsection{Scene Identification}
\label{subsec:scenes}

\begin{figure}[t!]
    \centering
    \includegraphics[width=0.48\textwidth]{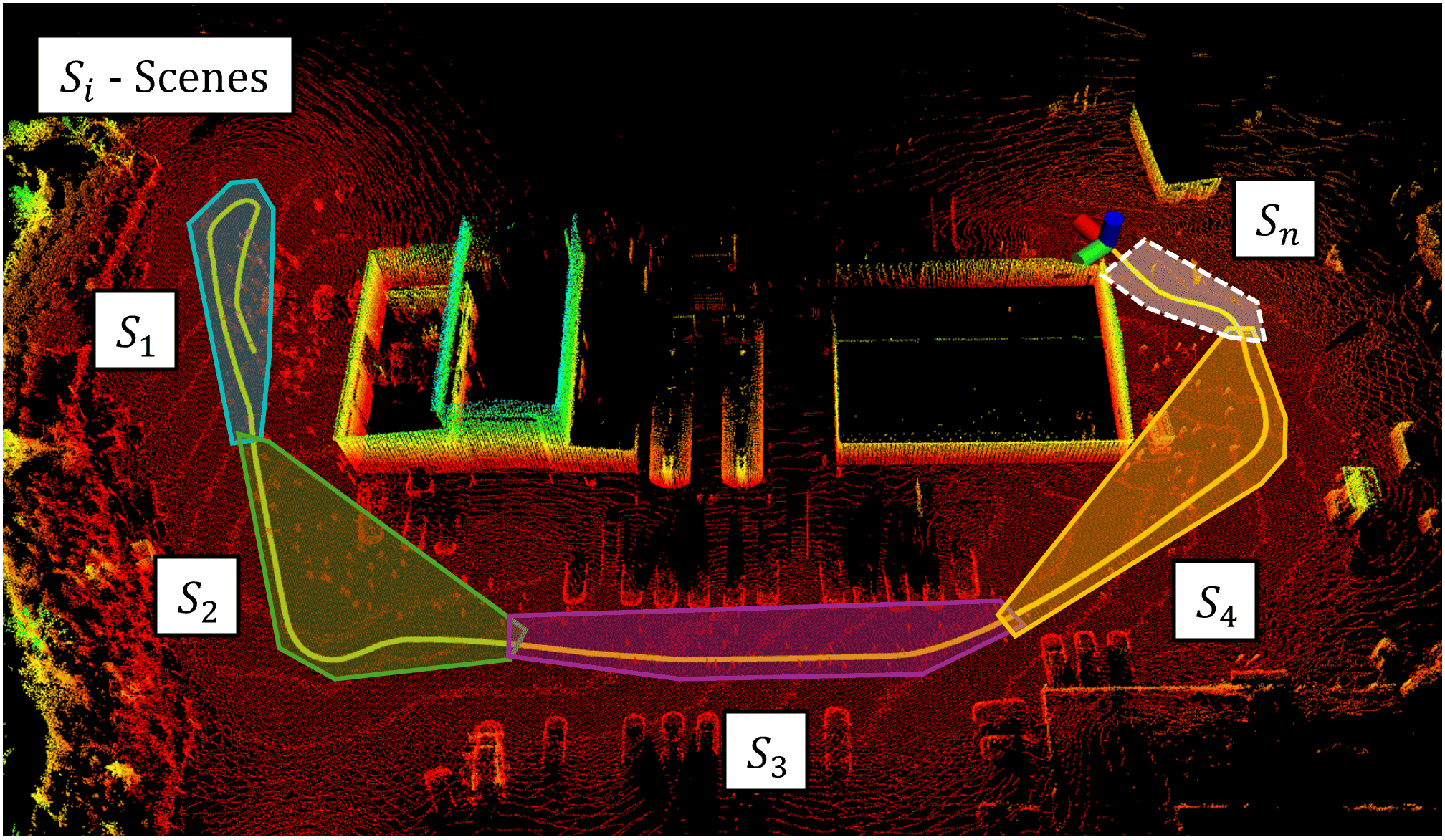}
    \caption{Scenes spatially decompose the sensor trajectory into convex hulls with minimal overlap. Each convex shape (denoted $S_{i}$) represents a scene where change detection occurs separately. The scene outlined in dashed white represents the current scene which is still being built.}
    \label{fig:scenes}
    \vskip -0.2in
\end{figure}

Here, we address the problem of partitioning the mapping session into scenes.
As the fundamental unit of the framework, each scene is a bounded region of the sensor trajectory that, together with the timestamps of each visit, defines a change detection instance (\cref{fig:scenes}).
Ideally, scenes are selected such that running change detection independently on every scene reports the same changes as one comparison over the complete map at a fraction of the cost.
Identifying suitable scene boundaries is therefore important because it impacts both computation time and detection accuracy.

Several factors that balance change detection accuracy and computation time must be considered when defining a good scene.
First, every scan must belong to at least one scene, i.e., completeness.
Second, scenes must be large enough to enable meaningful comparisons between $\mathcal{C}_{j}^{1:v-1}$ and $\mathcal{S}_{j}^{v}$.
Scenes that are too small degenerate into scan-to-scan comparisons, which inherently struggle to distinguish real changes from differences in coverage and viewpoint.
Third, to enable online computation times, scenes must be small enough for a fixed-size submap to sufficiently represent them.
Fourth, all scans within a scene must be spatially cohesive by exhibiting sufficient pairwise overlap.
This is to prevent wasteful cross-region comparisons during change detection instances.
We propose a greedy heuristic that constructs scenes incrementally as scans are collected, satisfying the completeness and size constraints.
Formulating optimal conditions for scenes and the online optimization of scenes is left to future work.

\begin{algorithm}[t]
\small
\caption{Online Incremental Scene Identification}
\label{alg:scene_identification}
\begin{algorithmic}[1]
    \renewcommand{\algorithmicrequire}{\textbf{Given:}}
    \renewcommand{\algorithmicensure}{\textbf{Output:}}
    \REQUIRE Scan Packet $\{k,P_{k}, \mathbf{x}_{k}, \mathbf{d}_{k}\}$
    \IF {$k=1$}
        \STATE $S_{cur} = \emptyset$
        \STATE Return
    \ELSIF {$S_{cur} = \emptyset$ and $\exists S \in \mathfrak{S}$ such that $\mathbf{x}_{k} \in S$}
        \STATE Return
    \ENDIF
    \STATE $S_{n} = \operatorname{conv}\bigl(S_{cur} \cup \{\mathbf{x}_{k}\}\bigr)$
    \IF {$|S_{n}| > \alpha$ or $\max_{\mathbf{d} \in D_{n}}||\mathbf{d}_{k}-\mathbf{d}||_{2} > \beta$}
        \STATE $\mathfrak{S}.append(S_{cur})$
        \STATE {\tt\footnotesize\color{blue}// Do not start new scene while in old scene}
        \IF {$\exists S\in \mathfrak{S} \text{ such that } \mathbf{x}_{k} \in S$}
            \STATE $S_{cur} = \emptyset$
        \ELSE   
            \STATE $S_{cur} = \mathbf{x}_{k}$
        \ENDIF
    \ELSE
        \STATE $S_{cur} = S_{n}$
    \ENDIF
\end{algorithmic}
\end{algorithm}

\Cref{alg:scene_identification} gives the procedure for constructing scenes incrementally.
The algorithm receives a packet with each scan containing the sequence number $k$, point cloud $P_{k}$, position $\mathbf{x}_{k}$, and descriptor $\mathbf{d}_{k}$.
The algorithm initializes an empty active scene the first time it is called (lines 1-3).
When a packet is received and there is an active scene, the algorithm checks if the scene can expand to include the new scan or if the active scene should be closed.
Line 7 shows the scene expansion step, where a potential new scene $S_{n}$ is found by taking the convex hull of the active scene and the new pose $\mathbf{x}_{k}$.
Line 8 checks if this new scene violates the size constraint set by a parameter $\alpha$ or a scan pairwise overlap condition set by parameter $\beta$, where we use the shorthand notation $D_{n}$ to denote the descriptor set in $S_{n}$.
Recall that the distance between descriptors monotonically decreases with the degree of overlap between scans, so the pairwise overlap condition ensures all scans in a scene have sufficient overlap.
If either constraint is violated, the algorithm will attempt to start a new scene.
However, to avoid wasted change detection instances caused by overlapping scenes, a new scene is opened only if the current pose does not already fall within a previously established scene (lines 4 \& 11).

Entry and exit timestamps are recorded for each visit to a scene.
For a scene that has been visited $v$ times, the exit timestamp of the $(v-1)$ visit is $t_{out}^{v-1}$, and the entry timestamp of the $v$ visit is $t_{in}^{v}$.
The scene position bounds and these timestamps fully define a change detection instance.

\subsection{Submap Generation}
\label{subsec:submap_gen}

We now describe the process for selecting submaps that accurately represent a scene as defined by the physical and temporal constraints outlined in the previous subsection.
Our approach uses OptMap \cite{thorne2025optmap} to select size-constrained, maximally informative submaps for each scene. 
OptMap operates over scan descriptors rather than the point clouds, so submap selection reduces to computing vector norms, and the only points loaded are relevant to the current change detection instance. 
Using a prior global map instead requires a trade-off between local point density---which is required for accurate change detection---and memory allocation.
We first define submap selection, then summarize relevant results from \cite{thorne2025optmap}, and finally provide implementation details for this application.

First consider scene $S_{j}$ that has been visited $v$ times and has timestamps $t_{out}^{v-1}$ and $t_{in}^{v}$. 
The cumulative submap $\mathcal{C}_{j}^{1:v-1}$ will be a subset of scans in $S_{j}$ collected before $t_{out}^{v-1}$, and the recent submap $\mathcal{S}_{j}^{v}$ will be a subset of scans in $S_{j}$ collected after $t_{in}^{v}$.
These submaps are selected by optimizing over the corresponding sets of descriptors for the scene $D_{S}^{c} = \{\mathbf{d} \in \mathbb{R}^d \, | \, \forall (\mathbf{d}, \mathbf{x}, t) \in \mathfrak{T}, \, \mathbf{d} \in D_{S}, \, t \leq t_{out}^{v-1}\}$ and $D_{S}^{r} = \{\mathbf{d} \in \mathbb{R}^d \, | \, \forall (\mathbf{d}, \mathbf{x}, t) \in \mathfrak{T}, \, \mathbf{d} \in D_{S}, \, t \geq t_{in}^{v}\}$, for the cumulative and recent submaps respectively.

Thus the objective is to select two cardinality-constrained descriptor subsets $C \subseteq D_{S}^{c}$ and $R \subseteq D_{S}^{r}$ that each maximize $\Gamma(E) = L_{cebc}(\textbf{0}) - L_{cebc}(E \cup \textbf{0})$, where $\textbf{0}$ is a vector of zeros, $E$ is the solution descriptor set, and $L_{cebc}: \mathcal{D} \rightarrow \mathbb{R}$ is a clustering-based loss function defined as $L_{cebc}(E) = \frac{1}{w_{tot}}\sum_{\mathbf{d} \in D_{S}}w_{\mathbf{d}}\min_{e \in E}||\mathbf{d} - e||_{2}$.
When selecting a cumulative submap $D_{S} = D_{S}^{c}$ and when selecting a recent submap $D_{S} = D_{S}^{r}$.
$L_{cebc}$ is a weighted clustering function where each descriptor $\mathbf{d}$ is assigned a weight $w_{\mathbf{d}}$ equal to the Euclidean distance between $\mathbf{d}$ and the preceding descriptor in the mapping session.
The term $w_{tot}$ is a normalization term equal to the sum of weights $w_{tot} = \sum_{\mathbf{d} \in D_{S}} w_{\mathbf{d}}$ such that $L_{cebc} \in [0,1]$.
In \cite{thorne2025optmap}, $\Gamma(R)$ is proven to be submodular, and the OptMap algorithm is shown to solve it within a theoretical $1/2-\epsilon$ bound where $\epsilon$ is a granularity constant, and in practice most solutions are shown to significantly outperform the theoretical bound.

Once the descriptor subsets are selected, the corresponding point clouds are loaded where $\mathcal{C}_{j}^{1:v} = \{P_{k} | \forall(\mathbf{d}_{k},\mathbf{x}_{k},t_{k}) \in \mathfrak{T}, \mathbf{d}_{k} \in C\}$ and $\mathcal{S}_{j}^{v} = \{P_{k} | \forall(\mathbf{d}_{k},\mathbf{x}_{k},t_{k}) \in \mathfrak{T}, \mathbf{d}_{k} \in R\}$.
Notably, OptMap can gracefully handle scenes that have been visited three or more times, where $\mathcal{C}_{j}^{1:v-1}$ can be selected from two or more mapping passes.
Alternative methods for submap generation struggle to select a representative subset given large potential discontinuities in the input set.
The selected $\mathcal{C}_{j}^{1:v-1}$ and $\mathcal{S}_{j}^{v}$ are then used for the change detection algorithm presented in the following subsection.

\subsection{Change Detection using Submaps \& Occlusion Filtering}
\label{subsec:change_detect_alg}

\begin{figure}[t!]
    \centering
    \includegraphics[width=0.48\textwidth]{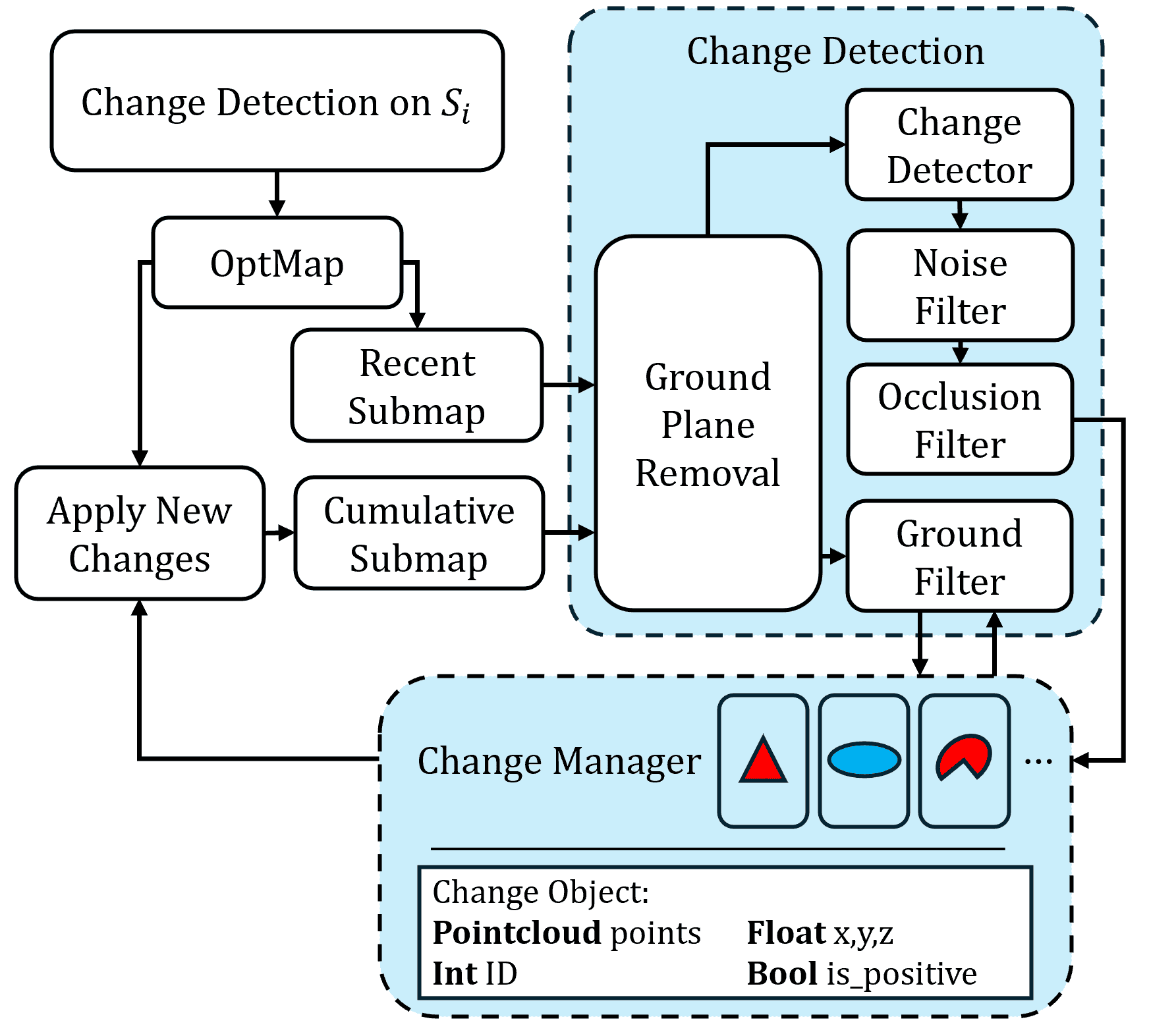}
    \caption{Before the change detection algorithm occurs, OptMap \cite{thorne2025optmap} is called to generate $\mathcal{C}_{j}^{1:v-1}$ and $\mathcal{S}_{j}^{v}$. During change detection five sequential steps occur: 1) ground plane removal (segmentation), 2) nearest-neighbor based change detection, 3) occlusion filtering, 4) noise filtering, and 5) ground plane filter after changes are clustered into candidate objects. Detected changes are clustered and managed as objects with respective change clouds, object IDs, average coordinates, and a positive/negative flag.}
    \label{fig:pipeline}
    \vskip -0.2in
\end{figure}

This section describes the change detection algorithm, which takes $\mathcal{C}_{j}^{1:v-1}$ and $\mathcal{S}_{j}^{v}$ from OptMap as inputs and returns a set of positive and negative change points.
We define change points as points that can be confirmed to appear, disappear, or move between passes---a stricter criterion than methods that label any unmatched point as a change.
This strict definition requires confirming that candidate change points have not simply been occluded due to differing fields of view between the submaps, motivating a novel occlusion filter.
We additionally include a ground segmentation step to make detecting objects near the ground easier, and a novel ground filter that prevents ground segmentation errors from propagating into false change detections.
The complete pipeline is shown in the change detection block of \cref{fig:pipeline}, and described in the following.

The first step is ground segmentation for $\mathcal{C}_{j}^{1:v-1}$ and $\mathcal{S}_{j}^{v}$ separately.
The approach follows Patchwork++ \cite{lee2022patchwork++}, dividing the submap into a polar grid and fitting a ground plane Gaussian distribution estimate with a mean position and covariance for each bin.
Points are labeled as non-ground if their Mahalanobis distance to the ground plane exceeds a threshold.
We assume the ground does not change between passes and therefore only perform change detection on non-ground points, though this assumption is not binding and change detection could be applied to ground points separately in certain applications.

The change detector follows ground segmentation using a nearest neighbor distance threshold.
A non-ground point is labeled as a \emph{candidate change} if the average distance to the $n$ nearest neighbors in the corresponding submap exceeds a threshold $\tau$.
As a rule of thumb, $\tau$ should be set greater than $d_{avg}/2$, where $d_{avg}$ is the larger of the average nearest-neighbor distances in $\mathcal{C}_{j}^{1:v-1}$ and $\mathcal{S}_{j}^{v}$.
This threshold effectively sets the minimum resolution of detectable change objects.
Following nearest neighbor thresholding comes a simple noise filter which requires the average distance from a change point to its $n$ nearest change points to be below a threshold $\rho$.

\begin{figure*}[t!]
    \centering
    \includegraphics[width=0.96\textwidth]{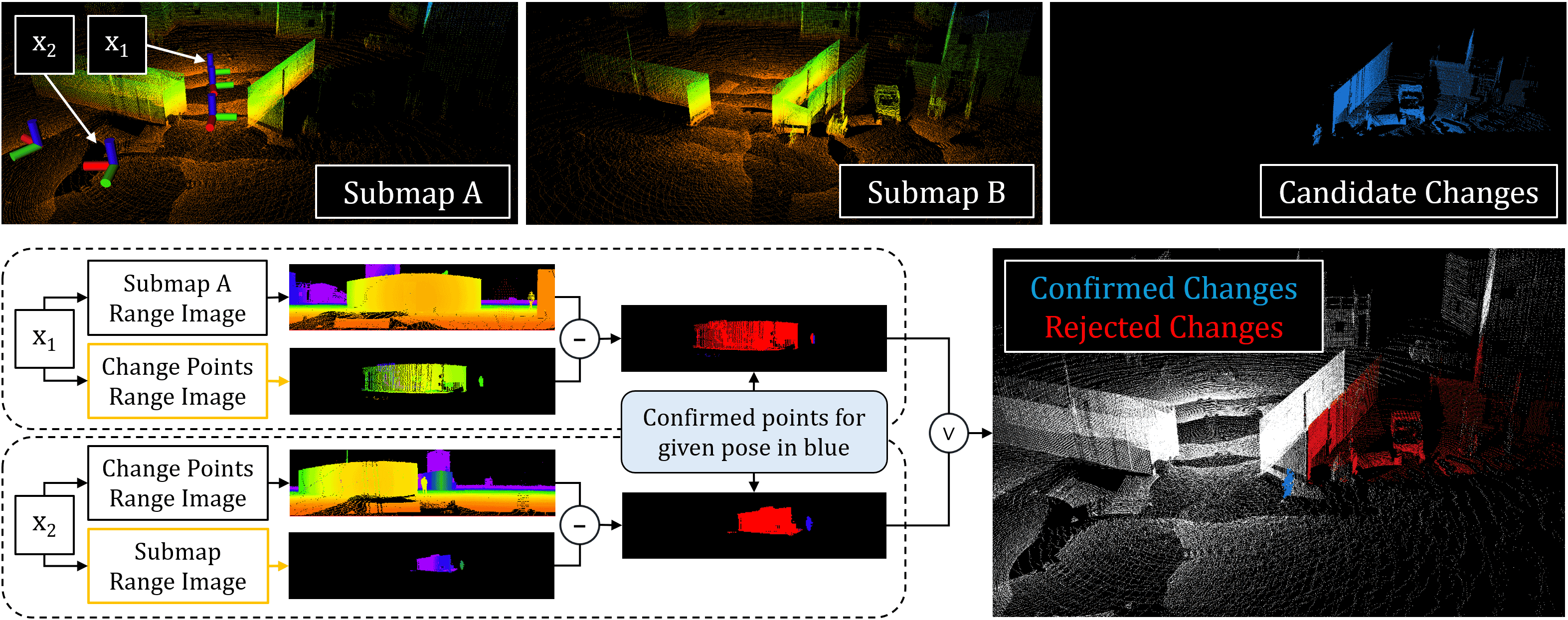}
    \caption{Occlusion filter on a submap-to-submap change detection instance. Top three images show submap A, B, and the set of candidate change points that are in submap B but not A. Bottom-left shows range image projection and subtraction for submap A and candidate change points at each scan pose in submap A. If a point is confirmed by any range image subtraction step it is added to the set of confirmed changes shown in the bottom-right.}
    \label{fig:occlusion_filter}
    \vskip -0.2in
\end{figure*}

Candidate change points are then passed through an occlusion filter.
This step is necessary because the sensor's trajectory will likely be different from the one taken when the scene was created, meaning the submaps might contain points from different parts of the environment due to differing fields of view.
This intuition is shown in \cref{fig:occlusion_filter}: submap B observes a large area that submap A cannot because a large wall occludes it.
Our approach to confirming a candidate change point is to check whether points were collected behind the candidate change from the perspective of the opposing submap.
In other words, a positive change candidate from submap B must be confirmed by checking if submap A could have observed the change or not.
In \cref{fig:occlusion_filter}, the candidate changes were observed by submap B but not A, so the occlusion filter makes a set of range image pairs for each scan in submap A by projecting the candidate change points to the location of the scan and projecting submap A to the location of the scan.
These paired range images are visible in the bottom-left of \cref{fig:occlusion_filter}.
Pixel-wise subtraction between paired range images yields a new range image where the pixel value for a confirmed point is positive if and only if a point in submap A was observed behind the candidate change point.
This range image process simplifies what is normally performed as an expensive ray tracing step.

The final step is the ground filter, which is applied after candidate change points have been clustered (see \cref{subsec:change_manager}).
Ground segmentation failures (e.g., due to reflections from standing water) in $\mathcal{C}_{j}^{1:v-1}$ or $\mathcal{S}_{j}^{v}$ may result in incorrectly labeled changes.
The ground filter addresses this by treating each change object as a Gaussian distribution and comparing them to the nearest ground plane estimate. 
All positive change objects are compared to the ground estimate from $\mathcal{C}_{j}^{1:v-1}$, and negative change objects to the ground estimate from $\mathcal{S}_{j}^{v}$.
For a change object to be rejected, it must satisfy two separate conditions motivated by the Bhattacharyya distance for measuring the similarity between Gaussian distributions (\cite{kailath1967divergence}, Eq. 61).
First, the object must be close to the ground as measured by the signed distance from its mean position $m_{o}$ to the ground plane estimate's mean position $m_{g}$ and unit normal vector $n_{g}$: $B_{1} = |(m_{o}-m_{g})\cdot n_{g}|$.
Second, it must have a similar covariance according to the second term of the Bhattacharyya distance
\begin{equation*}
    B_{2} = \frac{1}{2} \log\left(\frac{\det (\frac{\Sigma_{g} + \Sigma_{o}}{2})}{\sqrt{\det\Sigma_{g}\cdot\det\Sigma_{o}}}\right),
\end{equation*}
where $\Sigma_{g}$ is the covariance of the ground and $\Sigma_{o}$ is the covariance of the change object.
Given parameters $c_{1}$ and $c_{2}$, a change object is rejected if both $B_{1}<c_{1}$ and $B_{2}<c_{2}$.

\subsection{Change Manager}
\label{subsec:change_manager}

\begin{table*}[t]
    \small
    \centering
    \renewcommand{\arraystretch}{0.95}
    \caption{Change-detection ablation (mean of 5 runs per variant), reported as the average across the single-session Short (4 changes) and Long (18 changes) datasets. A ground-truth object counts as detected when a change cluster of the same polarity falls within 1\,m of it. GT detection percentages are cumulative: \emph{Whole} counts objects with $>\!75\%$ of the GT object's points, \emph{Partial} counts those at $>\!50\%$. Spurious detections are split into small and large at 125 points. \emph{Dropped} is the mean number of change-detection instances discarded because of a full queue. \textbf{Bold} marks the best value, \underline{underline} marks the second best value, and \textcolor{red}{red} marks the worst value in each column. Middle group: alternative detection frameworks. Bottom group: ablations of our framework.}
    \label{tab:results}
    \begin{tabular*}{\textwidth}{@{\extracolsep{\fill}} l cc cc cc c @{}}
    & \multicolumn{2}{c}{GT detected (\%)} & \multicolumn{2}{c}{Spurious} & \multicolumn{2}{c}{Compute (s)} & \\
    \cmidrule(lr){2-3} \cmidrule(lr){4-5} \cmidrule(lr){6-7}
    Method
    & \makecell{Whole\\($>$75\%)} & \makecell{Partial\\($>$50\%)}
    & \makecell{Small\\($<$125 pts)} & \makecell{Large\\($\geq$125 pts)}
    & Inst.\,$\downarrow$ & Sess.\,$\downarrow$ & \makecell{Dropped\\$\downarrow$} \\
    \midrule
    CDSD (Ours)       & 89.4          & \textbf{97.8} & \textbf{2.1}     & $\underline{2.0}$ & 1.51              & $\underline{41.8}$ & 0   \\
    \midrule\midrule
    Scan-to-submap           & 29.4          & 53.6          & 3.4              & 2.2              & \textbf{0.65}     & 68.9              & 26  \\
    Scan-to-map              & \textcolor{red}{10.6}          & \textcolor{red}{16.1}          & 6.2              & 34.1             & \textcolor{red}{2.32}              & \textcolor{red}{100.9}             & \textcolor{red}{103} \\
    \midrule\midrule
    w/o OptMap Submap     & 52.2          & 79.7          & 3.4              & \textbf{1.0}     & $\underline{1.20}$ & \textbf{31.9}     & 0   \\
    w/o Ground Removal       & 62.2          & 74.7          & $\underline{2.6}$ & 2.5              & 1.61              & 44.9              & 0   \\
    w/o Occlusion Filter     & $\underline{93.9}$ & 93.9      & \textcolor{red}{130.4}            & \textcolor{red}{282.2}            & 2.08              & 57.5              & 0   \\
    w/o Noise Filter         & 88.3          & $\underline{95.6}$ & 18.9         & 6.5              & 1.71              & 45.5              & 0   \\
    w/o Ground-Plane Filter  & \textbf{95.6} & \textbf{97.8} & 16.1             & 35.0             & 1.55              & 43.8              & 0   \\
    \bottomrule
    \end{tabular*}
    \vskip -0.1in
\end{table*}


\begin{table}[t]
    \small
    \centering
    \renewcommand{\arraystretch}{0.95}
    \caption{Sensitivity to the change detection distance gate $\tau$. All other parameters at their default values. The $\tau=0.30$ row is the default operating point. Results averaged over five runs on the Short and Long datasets.}
    \label{tab:thresh_sweep}
    \begin{tabular*}{\columnwidth}{@{\extracolsep{\fill}} l ccc @{}}
    \toprule
    \makecell{$\tau$\\(m)} & \makecell{GT whole\\($>$75\%)} & \makecell{Spurious\\Small ($<$125 pts)} & \makecell{Spurious\\Large ($\geq$125 pts)} \\
    \midrule
    0.1 & 88.9 & 8.6 & 8.4 \\
    0.2 & 83.1 & 1.5 & 3.4 \\
    0.3 & 89.4 & 2.1 & 2.0 \\
    0.4 & 88.1 & 1.7 & 1.2 \\
    0.5 & 71.1 & 1.4 & 0.8 \\
    \bottomrule
    \end{tabular*}
    \vskip -0.1in
\end{table}

\begin{table}[t]
    \small
    \centering
    \renewcommand{\arraystretch}{0.95}
    \caption{Sensitivity to the maximum scene volume $\alpha$. Scenes were restricted to 2D for these experiments, so $\alpha$ is reported as an area. All other parameters at their default values. The $\alpha=120$ row is the default operating point. Results averaged over five runs on the Short and Long datasets. Submap size increases proportionally with $\alpha$.}
    \label{tab:scene_sweep}
    \begin{tabular*}{\columnwidth}{@{\extracolsep{\fill}} l ccc @{}}
    \toprule
    \makecell{$\alpha$\\(m$^2$)} & \makecell{GT whole\\($>$75\%)} & \makecell{Spurious\\($\geq$50 pts)} & \makecell{Compute\\(s/inst.)} \\
    \midrule
    30  & 50.0 & 4.4 & 0.667 \\
    60  & 34.7 & 4.1 & 0.861 \\
    120 & 89.4 & 4.8 & 1.510 \\
    240 & 88.9 & 5.5 & 2.935 \\
    480 & 97.2 & 6.8 & 5.431 \\
    \bottomrule
    \end{tabular*}
    \vskip -0.2in
\end{table}

The change manager converts confirmed change points into higher-level abstractions via distance-threshold clustering.
Abstracting change points into objects enables efficient online map maintenance and makes detected changes actionable for downstream planners, which can query object-level information about the environment. 
For map maintenance, observations of scenes visited more than twice can be consolidated into a single $\mathcal{C}_{j}^{1:v-1}$ by applying all previously detected changes.
Each change object, shown by \cref{fig:pipeline}, contains the clustered points, a unique ID, a timestamp, the object's average coordinate, and a boolean indicating whether it is a positive or negative change.
The change manager provides the following functions for downstream use: \texttt{update\_map(cloud points, float time)} generates $\mathcal{C}_{j}^{1:v-1}$ by sequentially applying all negative change objects with timestamps less than the query time (only negative changes are needed since the input submap already contains all positive changes up to that time); and \texttt{publish\_change\_list(int n)} outputs metadata for all change objects with at least \texttt{n} points, giving planners a clean, noise-filtered view of detected changes.

%% file: sections/results.tex
\section{Hardware Platform}
\label{sec:hardware_results}

\begin{figure}[t!]
    \centering
    \includegraphics[width=0.48\textwidth]{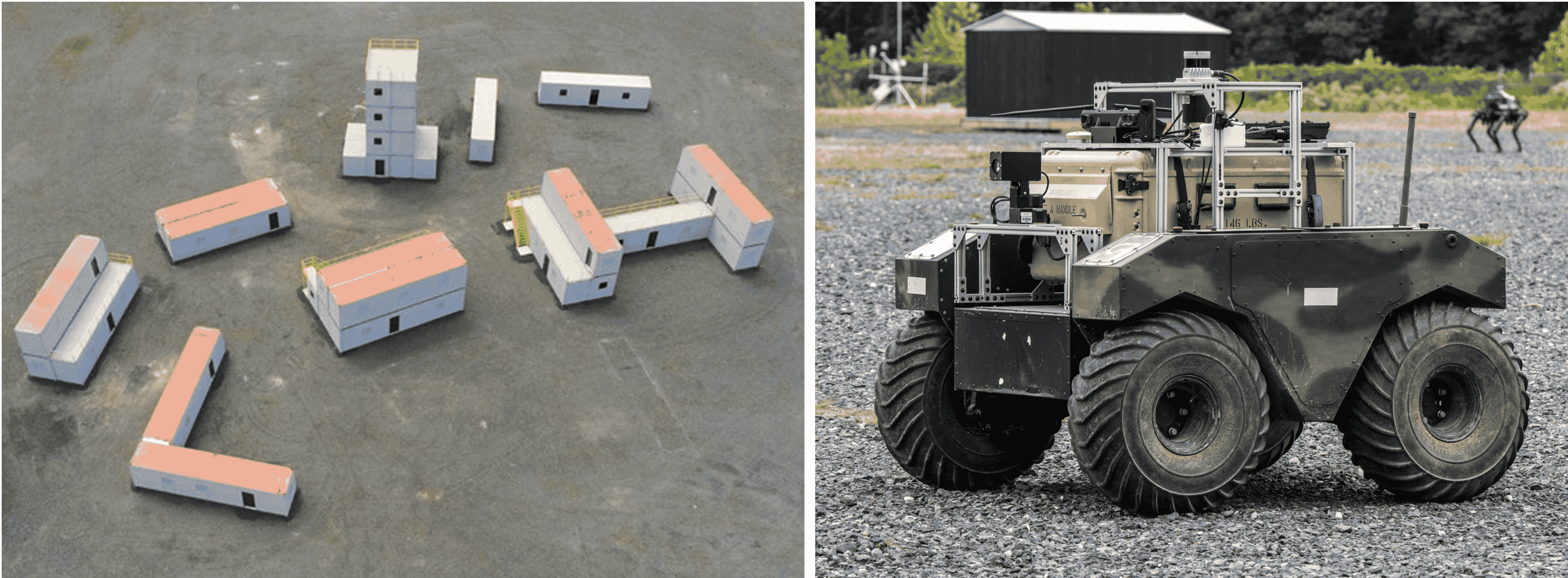}
    \caption{Left: \xblackout{Graces Quarters, MD} testing site. Right: Modified Clearpath Warthog with Ouster OS1 128-beam LiDAR used for experiments.}
    \label{fig:hardware}
    \vskip -0.2in
\end{figure}

Custom datasets were collected outdoors at the \xblackout{Army Research Laboratory facility in Graces Quarters, Maryland}.
The facility and robot equipped with an Ouster OS1 128-beam LiDAR are shown in \cref{fig:hardware}.
Two experiments were run with preplanned change objects moved during a single-session patrol mission, where the datasets collected are referred to as Short Patrol and Long Patrol or simply Short and Long.
The Short Patrol contains four preplanned changes and the Long Patrol contains eighteen.
The ground truth changes for the Short dataset are shown in \cref{fig:mout-change}.

\section{Results}
\label{sec:results}

We validate the proposed change detection framework with 1) an ablation study that demonstrates the value of each component of the framework, and 2) an experiment that compares detected changes on an open-source dataset to a state-of-the-art offline LiDAR change detection algorithm \cite{gil2025ephemerality}.
The results are generated with real-time sensor playback on a laptop i7-13620H CPU and NVIDIA 4060 GPU for generating descriptors.
In addition to the custom datasets described in \cref{sec:hardware_results}, the open-source dataset is the parking lot dataset originally presented in \cite{kim2022lt}, collected with an Ouster OS1 64-beam LiDAR.
Default parameters used for this testing are available alongside the open-source code.

\subsection{Ablation Study}

As noted by \cite{kim2022lt}, no open-source point-wise ground truth change detection datasets are available in the form of raw sensor data, so we chose to validate our approach with an ablation study on our Short and Long Patrol datasets.
Instead of looking to create a point-wise ground truth for the custom datasets, we hand label the position of each known change object and determine the approximate number of points in the abstracted object by running an offline variant of change detection.
To classify a ground-truth change object as detected, a change cluster of the same polarity (positive or negative) must fall within 1 m of it and reach at least 50\% of the object's point count, with matches above 50\% deemed partial and matches above 75\% deemed complete.

\begin{figure}[t!]
    \centering
    \includegraphics[width=0.48\textwidth]{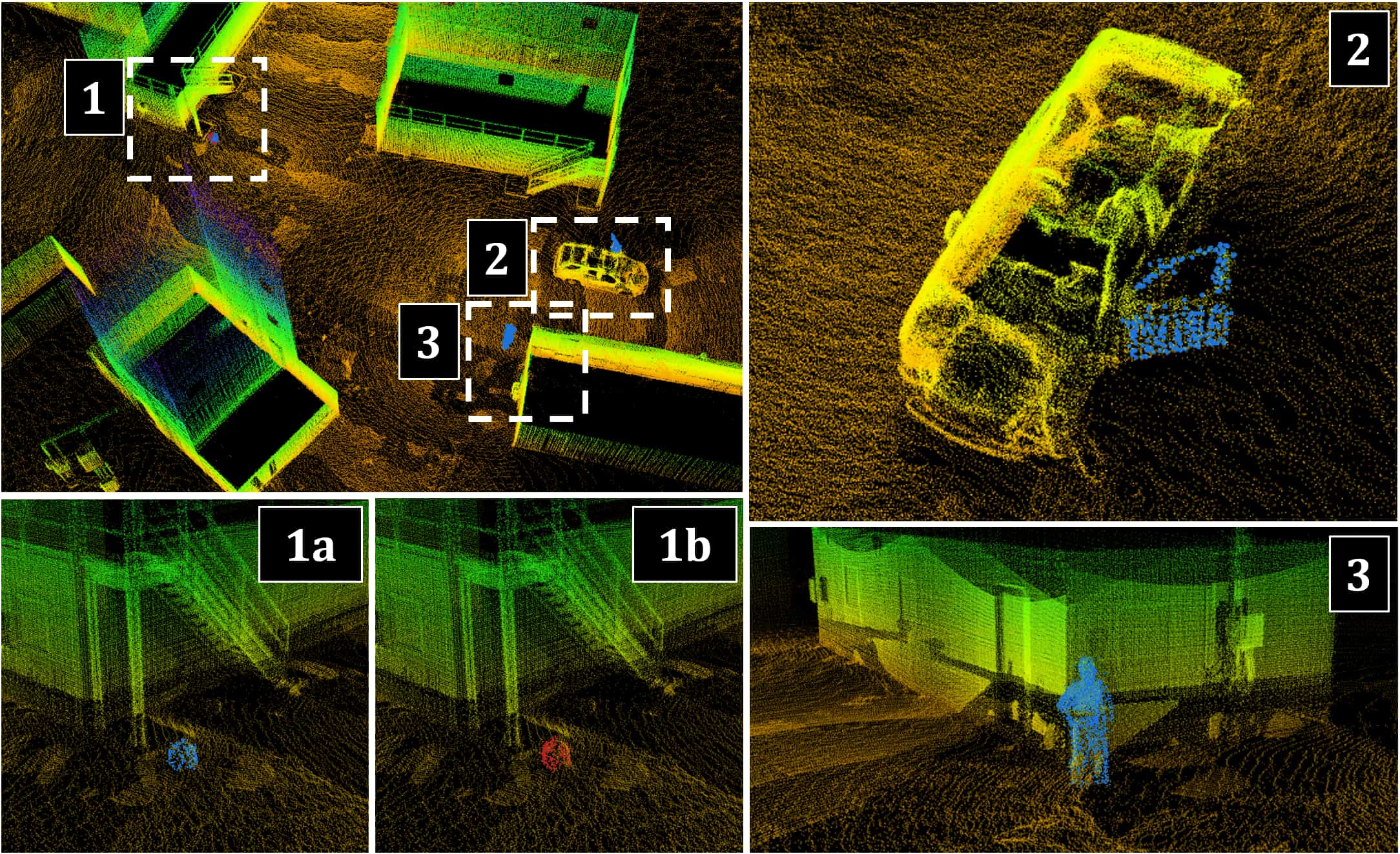}
    \caption{Ground truth changes for the Short change detection dataset. Top-left shows an overview of the map and views of each of the four changes. View 1 shows a backpack ($\sim20$ cm) which was introduced for view 1a and removed for view 1b. View 2 shows a car with a door opened between observations, and view 3 shows a person who appeared between observations.}
    \label{fig:mout-change}
    \vskip -0.25in
\end{figure}

\begin{figure*}[t!]
    \centering
    \includegraphics[width=1.0\textwidth]{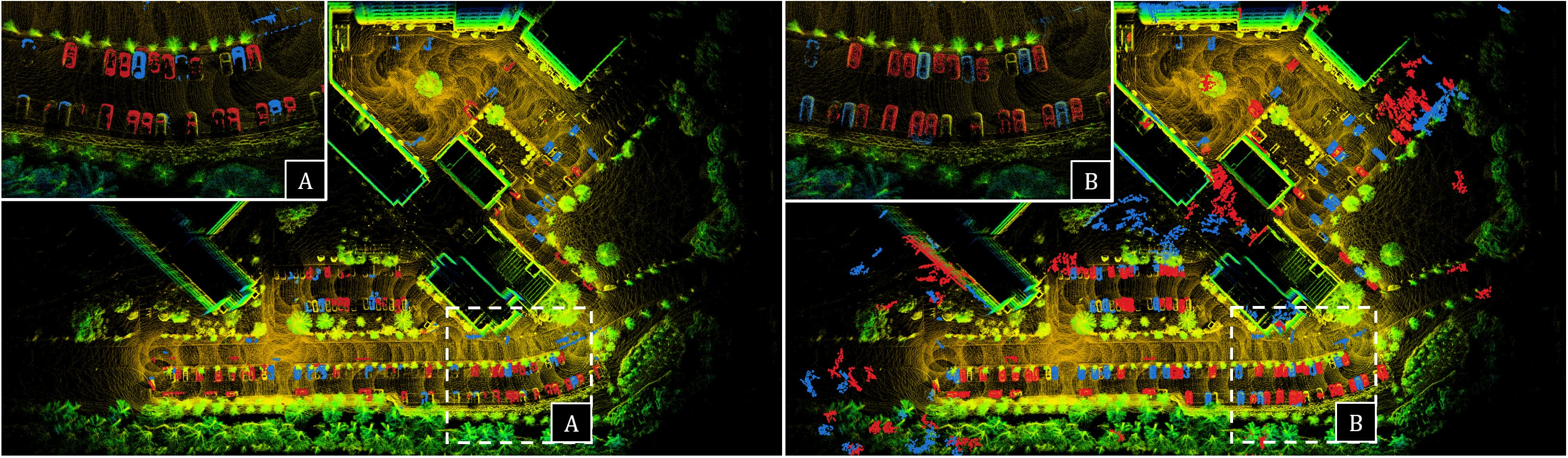}
    \caption{Change detection results from LT-Mapper Parking Lot dataset session 02 (Jan 25, 2021) to session 04 (Jan 26, 2021). Left figure shows the changes from CDSD, and right figure shows the changes detected by ELite \cite{gil2025ephemerality}. Positive change points are colored blue, negative change points are colored red, and the background map is colored by z-axis. The inset on the top-left of both figures shows a detailed difference in detection outputs for either algorithm.}
    \label{fig:parking_lot}
    \vskip -0.2in
\end{figure*}

\Cref{tab:results} compares CDSD against two alternative online change detection strategies and an ablation comparing each component of the proposed framework.
The two alternate strategies---Scan-to-Submap (S2S) and Scan-to-Map (S2M)---are inspired by other online geometric change detection algorithms \cite{wellhausen2017reliable, qian2022pocd}.
For both methods, a scan every 2.5 m traveled is compared to a prior map using the same set of filters that our method uses, where S2S uses scene-based OptMap submaps and S2M uses a dense keyframe prior map collected from the first pass of each dataset.
The ablation variant without OptMap uses a submap with scans selected via an equal time stride heuristic, and the variant without ground removal also does not use the ground filter.
We evaluate each method based on the percent of known ground truth objects detected, number of spurious or incorrect changes detected, the per instance computation time, total computation time, and the number of dropped change detection instances.
The cutoff for small and large spurious changes is set at 125 points to approximately match the smallest ground-truth change across both datasets.

S2M reaches only 10.6\% whole and 16.1\% partial detection and is not comparable in detection quality to CDSD's 89.4\% whole detection rate.
S2S recovers 29.4\% of ground-truth objects whole and 53.6\% partially.
Per-instance computation time is 54\% longer for S2M and 57\% shorter for S2S, but on average S2M completed 43 change instances and S2S completed 106, requiring 141\% and 65\% more total computation time respectively overall despite each dropping many change instances.
It is worth noting that S2S outperformed S2M across every category, indicating that scene-based submaps are a broadly useful addition to geometric change detection.

The bottom group of \cref{tab:results} ablates the submap selection strategy and each change detection filter.
Replacing OptMap's submodular scan selection with equal-time-spaced scans drops whole-object detection from 89.4\% to 52.2\%.
This 37.2\% drop is due to the fact that OptMap generates better cumulative submaps than evenly spaced heuristics, especially when the change occurs after a scene has been visited more than twice.
Removing ground segmentation reduces complete detections to 62.2\%, a decline exemplified by the backpack changes in \cref{fig:mout-change}, which are detected only 10\% of the time without ground segmentation.
Removing the occlusion filter increases whole-object detection rates to 93.9\%, at the cost of 130.4 small and 282.2 large spurious detections on average per session.
Removing either the noise or ground-plane filter similarly increases spurious detections (18.9/6.5 and 16.1/35.0 small/large when removed, versus 2.1/2.0 for CDSD).
Notably, per-instance computation time for CDSD is lower than every ablation except removing OptMap's submap selection (1.20\,s versus 1.51\,s).
Averaged across all runs, the top four most computationally expensive steps take 1.05\,s for nearest-neighbor search, 0.23\,s for ground segmentation, 0.16\,s for OptMap, and 0.09\,s for object clustering.
Aside from ground segmentation, each filter is computationally cheap to run and reduces the number of points that must be clustered.
Removing the occlusion, noise, and ground filter thus leads to between 2.6\% and 37.7\% increases in computation time.
Ground removal effectively offsets its required computation time by pruning points for the nearest-neighbor search which requires 0.34\,s longer to complete without ground removal.
In summary, each step is computationally efficient and meaningfully improves CDSD's detection performance.

\Cref{tab:thresh_sweep} and \cref{tab:scene_sweep} show sensitivity analysis for the minimum detection distance threshold $\tau$ and the scene size $\alpha$.
Both sweeps were averaged over five runs on the same two custom datasets used for \cref{tab:results}.
\Cref{tab:thresh_sweep} shows that low detection thresholds all detect the ground truth changes, but the lowest thresholds also detect many spurious changes, with $\tau=0.1$\,m producing by far the most (8.6 small, 8.4 large).
From $\tau=0.2$\,m upward, large spurious detections decrease monotonically (3.4, 2.0, 1.2, 0.8) while whole-object detection stays high until it falls off at $\tau=0.5$\,m (71.1\%).
The default value of $\tau=0.3$\,m catches almost all ground truth changes (89.4\% whole) with few spurious changes detected (2.1 small, 2.0 large).
\Cref{tab:scene_sweep} shows that whole-object detection is poor for the smallest scenes (50.0\% at $\alpha=30$\,m$^2$ and 34.7\% at $\alpha=60$\,m$^2$) but improves substantially as the scene grows (89.4\%, 88.9\%, and 97.2\% at $\alpha=120$, 240, and 480\,m$^2$ respectively), at the cost of per-instance compute time that grows from 0.667\,s to 5.431\,s across the sweep.
We note that the number of scans in each submap changed proportional to $\alpha$ and that $\alpha$ is reported as an area because scenes were restricted to 2D.
This matches the intuition that larger scenes lead to offline-like change detection performance in the limit, with $\alpha=120$\,m$^2$, the framework's default, offering a balance between accuracy and compute cost.

\begin{table}[t]
    \small
    \centering
    \caption{Corroboration of our online change detections by ELite~\cite{gil2025ephemerality} on the multi-session parking lot dataset~\cite{kim2022lt}. Total agreement is the percentage of all our change points that have an ELite change point of the same polarity within $0.5$\,m. The earlier session of each pair is used as the base map.}
    \label{tab:elite_total_precision}
    \setlength{\tabcolsep}{4pt}
    \begin{tabular*}{\columnwidth}{@{\extracolsep{\fill}} l c c @{}}
    \toprule
    Scenario & Sessions & Total Agreement (\%) \\
    \midrule
    Filling 1  & 02--03 & 89.6 \\
    Filling 2  & 04--06 & 89.1 \\
    Emptying 1 & 02--04 & 89.5 \\
    Emptying 2 & 03--04 & 89.4 \\
    \bottomrule
    \end{tabular*}
    \vspace{-0.2in}
\end{table}

\subsection{LT-Mapper Parking Lot Dataset}

We next quantify the agreement of the proposed method with ELite \cite{gil2025ephemerality}, a state-of-the-art open-source algorithm for offline LiDAR change detection, on the open-source multi-session LT-Mapper parking lot dataset \cite{kim2022lt}.
The four scenarios listed in \cref{tab:elite_total_precision} come from four sessions, two daytime (03 and 06) and two nighttime (02 and 04). 
They are named Filling 1--2 and Emptying 1--2 after the dominant direction of change in each, positive and negative respectively.
In a pre-processing step, we remove highly dynamic points using Removert \cite{kim2020remove} for both our framework and ELite following \cite{kim2022lt}, and otherwise use ELite's baseline approach and parameters.
We use PinNet \cite{ma2026pinnet} to align both sessions into a common coordinate frame so that our framework can detect multi-session changes and, for a fair comparison, trigger change detection only between sessions.
Agreement with ELite is measured with one-way point overlaps within $0.5$\,m.

The comparison is summarized by \cref{tab:elite_total_precision}, where 89.1\% to 89.6\% of our detected changes are corroborated by ELite.
The agreement is remarkably consistent between sessions indicating that our detections are reliably true changes.
The percent of ELite's detected changes in the session dominant polarity matched by our output is 54.2\%, 63.5\%, 50.5\%, and 55.7\% for Filling 1, Filling 2, Emptying 1, and Emptying 2 respectively.
We report ELite matches in the dominant polarity because ELite over-reports changes with respect to our change definition due to occlusion and unmatched fields of view.
Our method labels only changes that can be confirmed via the occlusion filter, whereas ELite treats any unmatched point in either map as a change---even when the mismatch simply reflects one session having mapped a portion of the parking lot that the other did not.
This is evident in \cref{fig:parking_lot}, where our results on the left come almost entirely from moved cars, while ELite reports numerous changes caused by field of view and position differences seen near the edges of the parking lot.
In short, our framework reports a conservative set of changes at high precision, reflecting a definition of change shaped by the assumptions and information available online rather than the complete-session view of an offline method.

%% file: sections/conclusion.tex
\section{Conclusion}
\label{sec:conclusion}

This paper presented CDSD, the first submap-based framework for online LiDAR change detection which spatially decomposes the environment into scenes to enable dense, accurate comparisons online.
Ablations confirmed that each of the four core components---scene identification, submap generation, occlusion-aware change detection, and change management---contribute to detection accuracy, and CDSD's online detections are consistently corroborated by a state-of-the-art offline method.
Future work can build semantic and temporal understandings of dynamic environments by combining strong geometric change detection with recent advances in local point descriptors and visual language models.